\documentclass[journal]{IEEEtran}

\ifCLASSINFOpdf
\else
   \usepackage[dvips]{graphicx}
\fi
\usepackage{url}

\usepackage{tabularx}
\usepackage{graphicx}
\usepackage{amsmath}
\usepackage{booktabs}
\usepackage{amssymb}
\usepackage{color}
\usepackage[cmyk]{xcolor}
\usepackage{balance}
\RequirePackage[caption=false,farskip=0pt,labelfont={}]{subfig}

\newcolumntype{Y}{>{\centering\arraybackslash}X}

\begin{document}

\title{Improving Domain Generalization on Gaze Estimation via Branch-out Auxiliary Regularization}

\author{Ruijie Zhao, \IEEEmembership{Student Member, IEEE}, Pinyan Tang, and Sihui Luo*, \IEEEmembership{Member, IEEE}
\thanks{This work was supported in part by the Natural Science Foundation Projects of Zhejiang Province [Grant NO. LQ22H090011 and LQ22F020020, 2021].}
\thanks{Ruijie Zhao, Pinyan Tang, and Sihui Luo (*corresponding author) are with the faculty of Electrical Engineering and Computer Science, Ningbo University, Ningbo, China (e-mail: 216001944@nbu.edu.cn, tangpinyan@nbu.edu.cn, luosihui@nbu.edu.cn).}}

\markboth{Submitted to IEEE signal processing letters}
{Shell \MakeLowercase{\textit{et al.}}: Bare Demo of IEEEtran.cls for IEEE Journals}
\maketitle

\begin{abstract}
Despite remarkable advancements, mainstream gaze estimation techniques, particularly appearance-based methods, often suffer from performance degradation in uncontrolled environments due to variations in illumination and individual facial attributes. Existing domain adaptation strategies, limited by their need for target domain samples, may fall short in real-world applications. This letter introduces Branch-out Auxiliary Regularization (BAR), an innovative method designed to boost gaze estimation’s generalization capabilities without requiring direct access to target domain data. Specifically, BAR integrates two auxiliary consistency regularization branches: one that uses augmented samples to counteract environmental variations, and another that aligns gaze directions with positive source domain samples to encourage the learning of consistent gaze features. These auxiliary pathways strengthen the core network and are integrated in a smooth, plug-and-play manner, facilitating easy adaptation to various other models. Comprehensive experimental evaluations on four cross-dataset tasks demonstrate the superiority of our approach.

\end{abstract}


\IEEEpeerreviewmaketitle

\section{Introduction}

\IEEEPARstart{G}{aze} estimation, which determines the gaze direction of an individual, has widespread applications across various domains such as autonomous driving \cite{liu2019gaze,shah2022driver}, virtual reality (VR) \cite{vr}, and psychological research \cite{rayner1998eye}. There are two primary approaches to gaze estimation:  model-based and appearance-based methods. Model-based methods \cite{yoo2005novel,zhu2005eye,zhu2006nonlinear,hennessey2006single,chen20083d,valenti2011combining} rely on expensive eye trackers and require detailed calibration for each user. On the other hand, appearance-based methods \cite{zhang2015appearance,fischer2018rt,kellnhofer2019gaze360,guan2023end,cheng2020gaze} leverage cost-effective webcams and deep learning models to infer gaze directions, offering a more accessible solution. 
\begin{figure}[t]
\centering
\includegraphics[width=\linewidth]{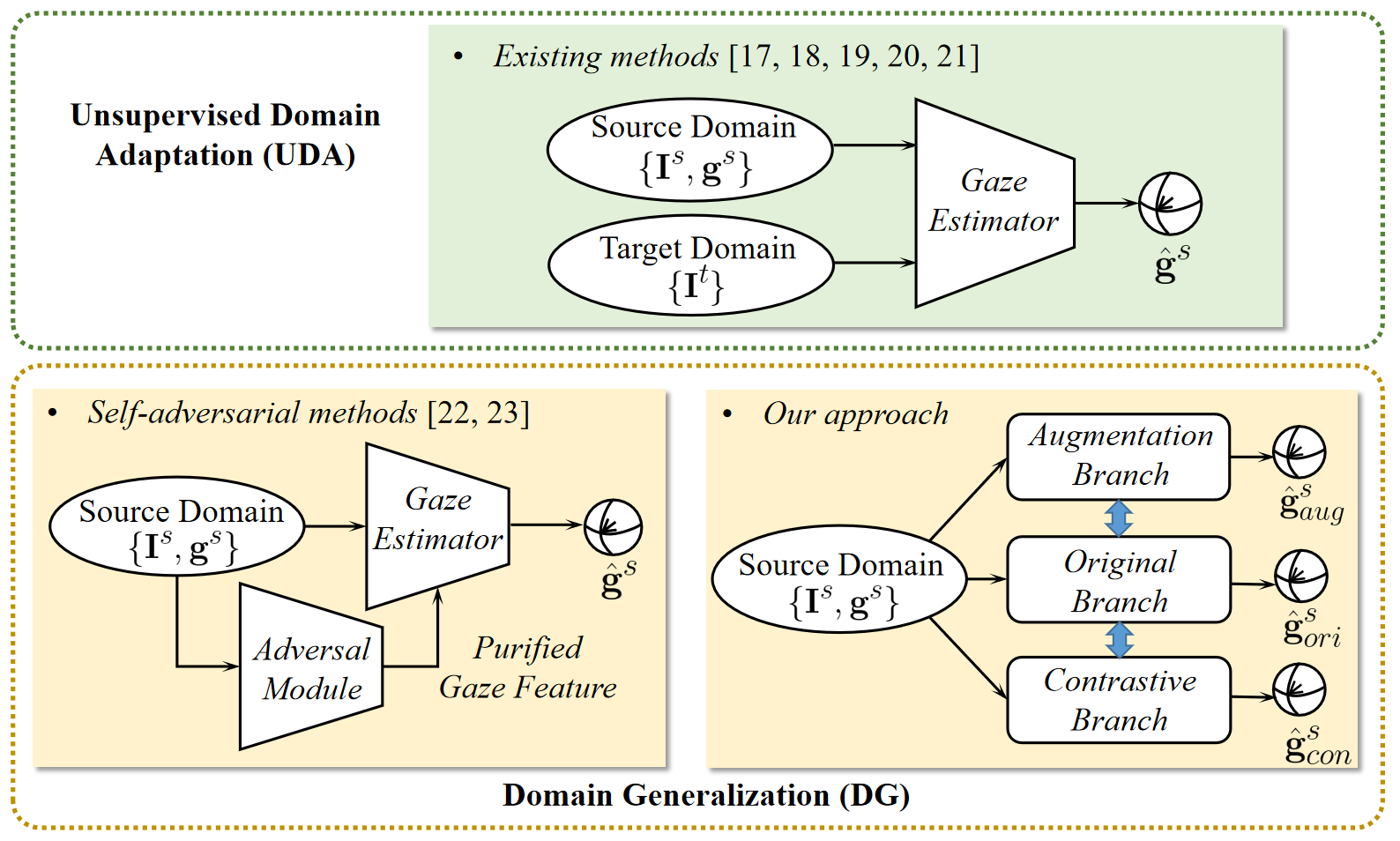}
\caption{Comparison of our method with conventional UDA and DG methods for gaze estimation. UDA methods generally rely on target domain data to enable knowledge transfer from the source domain to the target domain. Conventional DG methods leverage uncontrolled adversarial learning techniques that may cause feature elimination problems. Our method, in contrast, requires no access to target domain data and leverages flexible auxiliary consistency regularization branches to enhance model generalization.}
\label{fig:brief}
\end{figure}

Appearance-based methods demonstrate impressive performance within controlled environments \cite{zhang2017s}.
However, challenges arise in uncontrolled settings, where variations in illumination and diverse individuals pose obstacles to accurate gaze estimation, thereby hindering the generalization of gaze estimation models in cross-domain situations.

Recent studies \cite{wang2019generalizing,cai2023source,zhang2022outlier,wang2022contrastive,liu2024pnp} have explored the use of unsupervised domain adaptation techniques to extend gaze estimation models from source to target domains. However, these methods typically assume access to target domain samples, which poses challenges for practical application. Enhancing the generalization capabilities of gaze estimation models without direct access to the target domain presents significant hurdles. Recent research \cite{cheng2022puregaze, xu2023learning} employs adversarial learning to prevent models from learning gaze-irrelevant features, thereby mitigating domain-specific influences like illumination and identity. Nonetheless, the inherent complexity of gaze estimation and the risk of oversimplifying by removing irrelevant features might divert the model from its core aim of precise gaze estimation.

\begin{figure*}[t]
\centering
\includegraphics[width=14 cm]{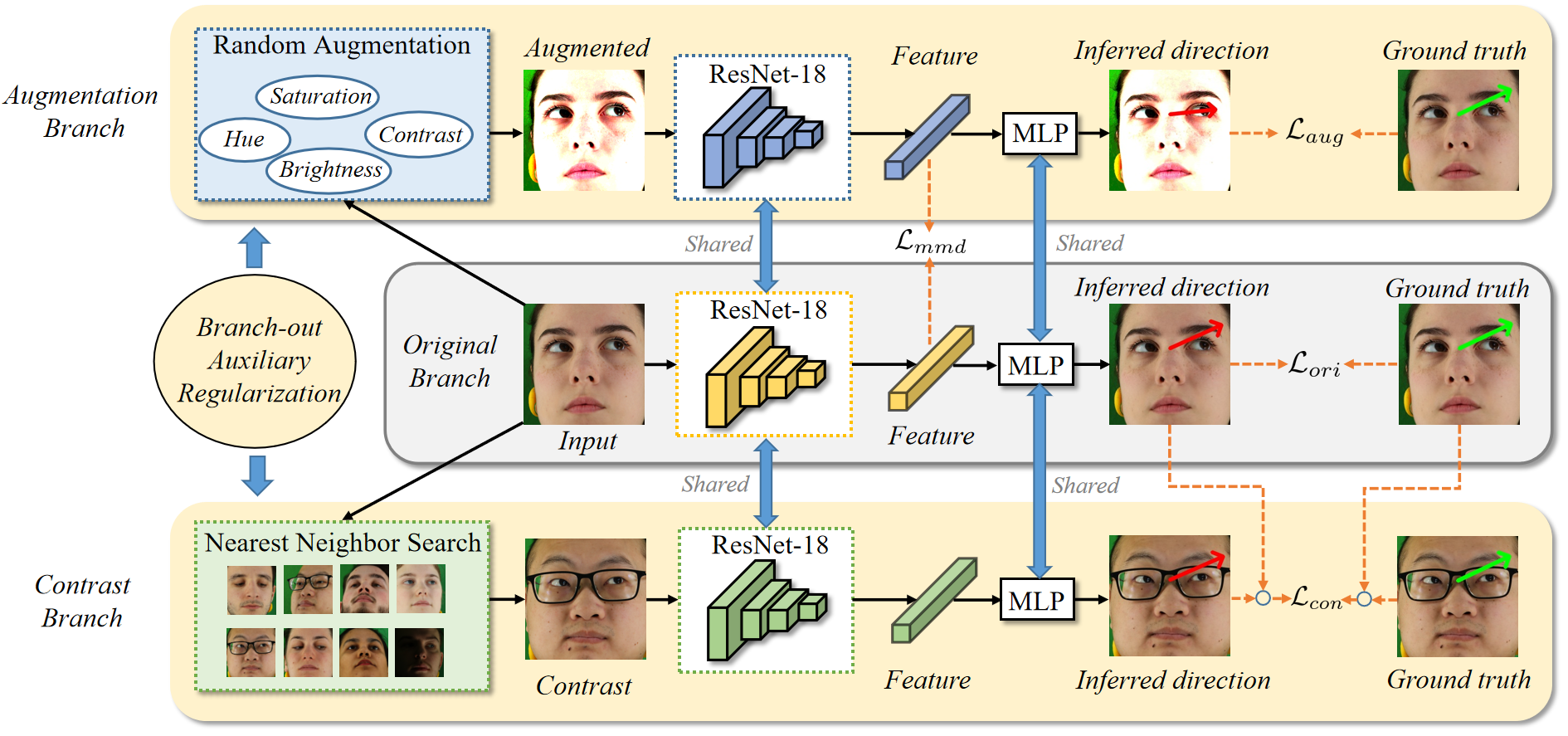}
\caption{The overall framework of our approach. We extend the original gaze estimation network by integrating two auxiliary consistency regularization pathways in a plug-and-play manner. Notably, the auxiliary branches are exclusively employed during the training phase and do not influence the test phase.}
\label{fig:dragon}
\end{figure*}

To address these challenges, we introduce Branch-out Auxiliary Regularization (BAR), a straightforward and flexible approach to enhance domain generalization in gaze estimation models. This framework enhances a pre-trained original gaze estimation network with two additional consistency regularization branches: the augmentation and the contrast branch. The augmentation branch inputs sophisticatedly augmented samples to reinforce the model’s generalization capability against environmental variations. Simultaneously, the contrast branch utilizes semantically consistent samples, which share similar gaze directions but differ in identity, to enforce the model to generate similar predictions. These enhancements enable the model to ignore gaze-irrelevant attributes and focus on learning gaze-consistent features, thereby avoiding the pitfalls of adversarial learning methods that may lead to the feature elimination problem. Fig.~\ref{fig:brief} illustrates the difference between conventional methods and our approach.

To summarize, the contributions of this letter are twofold. First, we present Gaze-BAR, a plug-and-play domain generalization method for gaze estimation that eliminates the need for target domain data and can be easily deployed to existing models to improve accuracy. Second, we propose to utilize auxiliary consistency regularization branches with sophisticated augmented samples and semantically consistent cross-identity samples as input to foster the model's generalization across varying environmental conditions and identities. Experimental results demonstrate that our method significantly surpasses the baseline model in performance and outperforms state-of-the-art domain generalization approaches for gaze estimation.

\section{Domain Generalization for Gaze Estimation}

\subsection{Preliminaries}

This section outlines the domain generalization task for gaze estimation. 
We begin by defining the source (training) domain as \(S_{train} = \{ (\mathbf{I}_i^{s}, \mathbf{g}_i^{s}) \}_{i=1}^{N_s}\), where \(\mathbf{I}_i^{s}\) represents the input image from the source domain, \(\mathbf{g}_i^{s}\) denotes the corresponding gaze direction, and \(N_s\) indicates the number of samples in the source domain dataset. Similarly, the target (testing) domain is denoted as \(S_{test} = \{ (\mathbf{I}_i^{t}, \mathbf{g}_i^{t}) \}_{i=1}^{N_t}\).

The formulation of the gaze estimation is represented as:
\begin{equation}
\hat{\mathbf{g}} = R(F(\mathbf{I})),
\end{equation}
where \(F(\cdot)\) represents the feature extractor, and \(R(\cdot)\) denotes the Multi-Layer Perceptron (MLP) to regress the feature vector into gaze direction. 

The predicted angular error $E[\mathbf{g},\hat{\mathbf{g}}]$ is defined as the angular disparity between the predicted gaze direction \(\hat{\mathbf{g}}\) and the actual gaze direction \(\mathbf{g}\): 
\begin{equation}
E[\mathbf{g}, \hat{\mathbf{g}}] = \arccos \left(\frac{\mathbf{g} \cdot \hat{\mathbf{g}}}{\|\mathbf{g}\| \|\hat{\mathbf{g}}\|}\right).
\end{equation}
Therefore, the objective is to optimize the model parameters \(\theta\) to
minimize the inference loss in the target domain: 
\begin{equation}
\theta = \arg \min_\theta \sum_i E[\mathbf{g}_i^t, \hat{\mathbf{g}}_i^t].
\end{equation}
With restricted access to the target domain, our approach aims to guide the model to focus on gaze-specific features while filtering out irrelevant ones. We identify two distinct categories of gaze-irrelevant factors:

\textbullet \ Environmental factors: Variations in image brightness, contrast, and tone lead to disparities.

\textbullet \ Facial attribute factors: Individual appearance variations such as skin complexion, facial morphology, and expressions.
 
To mitigate the effects of these gaze-irrelevant attributes, we introduce two auxiliary consistency regularization branches to the original model: the augmentation and the contrast branch. The augmentation branch encourages learning the consistent features regardless of varying environmental factors, while the contrast branch aims to disentangle gaze-specific features from irrelevant features concerning facial identity. The overall framework of our approach is shown in Fig.~\ref{fig:dragon}.

\subsection{Original Branch}
We employ ResNet-18~\cite{he2016deep} as the feature extractor backbone of the original gaze estimation model. The extracted feature vector is then passed through a single-layer MLP to regress the gaze direction. Besides, the original branch is trained with the L1 Loss between the predicted gaze direction $\hat{\mathbf{g}}_{ori}$ and the ground truth gaze direction ${\mathbf{g}}_{ori}$:
\begin{equation}
\mathcal{L}_{ori} = \vert \mathbf{g}_{ori} - \hat{\mathbf{g}}_{ori}\vert.
\end{equation}

\subsection{Augmentation Branch}
To address the challenge of varying illumination, we introduce the augmentation branch. Samples with random adjustments in brightness, contrast, and saturation are fed to this branch, simulating diverse environmental conditions to enhance the model’s resilience to such variations. This branch is trained with the L1 Loss \(\mathcal{L}_{aug}\), which is formulated as:
\begin{equation}
\mathcal{L}_{aug} = \vert\mathbf{g}_{ori} - \hat{\mathbf{g}}_{aug}\vert,
\end{equation}
where \(\hat{\mathbf{g}}_{aug}\) represents the predicted gaze direction for the augmented image.

Furthermore, to acquire consistent feature extraction across diverse illumination conditions, we align features from the original and augmented images by minimizing their disparity using Maximum Mean Discrepancy Loss (denoted as $\mathcal{L}_{mmd}$):
\begin{equation}
\mathcal{L}_{mmd} = \left\| \frac{1}{N} \sum_{i=1}^{N} \varphi
(x_i) - \frac{1}{N} \sum_{j=1}^{N} \varphi
(y_j) \right\|,
\end{equation}
where \( x_i \) and \( y_j \) denote the features extracted from the original and augmented images, \( \varphi \) represents the Gaussian kernel function, and \( N \) is the size of the feature. 
\subsection{Contrast Branch}

To enhance generalized gaze estimation, we tackle facial attribute variability through the contrast branch. This branch focuses on identifying positive samples with similar gaze directions but differing in facial attributes. These samples are selected from the source domain using the K-Nearest Neighbor (KNN) algorithm. For efficient searching, we employ the BallTree method \cite{dolatshah2015ball}, which expedites nearest neighbor searches by leveraging the query point’s location and its proximity to spherical regions, thus significantly cutting down search time.

In gaze estimation, $(pitch, yaw)$ denotes the vertical and horizontal angles of an individual’s gaze direction, with pitch and yaw referring to up-down and left-right eye movements, respectively.
Thus, we define the distance $d$ between two points in the searching tree as follows:

\begin{equation}
d = \sqrt{(pitch_2 - pitch_1)^2 + (yaw_2 - yaw_1)^2}.
\end{equation}

Here, (\(pitch_1, yaw_1\)) and (\(pitch_2, yaw_2\)) represent the coordinates of the query point and a point in the tree respectively.


The contrast branch aligns the gaze directions of positive samples with original images by minimizing contrast loss (\(\mathcal{L}_{con}\)), thereby directing the model to learn gaze-relevant features and disentangle gaze-irrelevant features concerning identity attributes. \(\mathcal{L}_{con}\) is computed as follows:
\begin{equation}
\mathcal{L}_{con} = \vert\hat{\mathbf{g}}_{con} - \hat{\mathbf{g}}_{ori}\vert,
\end{equation}
where \(\hat{\mathbf{g}}_{ori}\) and \(\hat{\mathbf{g}}_{con}\) represent the gaze predictions for the original image and the positive samples, respectively. However, due to the precision error stemming from the KNN search, disparities arise between the gaze direction of positive samples and the annotated direction of the original image. Therefore, we modify the \(\mathcal{L}_{con}\) as:
\begin{equation}
\mathcal{L}_{con} = \vert(\hat{\mathbf{g}}_{con} - \hat{\mathbf{g}}_{ori}) - (\mathbf{g}_{con} - \mathbf{g}_{ori})\vert,
\end{equation}
the subtraction term \((\mathbf{g}_{ori} - \mathbf{g}_{con})\) helps mitigate precision errors introduced by KNN.

\subsection{Overall Loss Function}
To unify the learning objectives of the original, augmentation, and contrast branches, we combine their respective loss functions into a single formulation. This holistic loss function, denoted as \(\mathcal{L}_{total}\), is formulated as:
\begin{equation}
\mathcal{L}_{total} = \mathcal{L}_{ori} + \lambda_{a}\mathcal{L}_{aug} + \lambda_{m}\mathcal{L}_{mmd} + \lambda_{c}\mathcal{L}_{con}.
\end{equation}

Here, $\lambda_{a},\lambda_{m},\lambda_{c}$ are hyperparameters and we set $\lambda_{a}=\lambda_{m}=\lambda_{c}=1.0$ in our implementation. 

\section{Experiment}

\begin{table*}[htbp]
\centering
\caption{Ablation studies of different components within our model}
\begin{tabularx}{\linewidth}{lYYYYY}
\toprule
Method & $\mathcal{D}_E\rightarrow\mathcal{D}_M (\downarrow)$& $\mathcal{D}_E\rightarrow\mathcal{D}_D(\downarrow)$ & $\mathcal{D}_G\rightarrow\mathcal{D}_M(\downarrow)$ & $\mathcal{D}_G\rightarrow\mathcal{D}_D (\downarrow)$& 
$Average (\downarrow)$\\
\midrule

$\mathcal{L}_{ori}$  
& $8.02^\circ\pm0.09^\circ$
&$ 9.11^\circ\pm0.31^\circ$
& $8.04^\circ\pm0.06^\circ$
& $9.20^\circ\pm0.08^\circ$
& $8.59^\circ$
\\
$\mathcal{L}_{ori} + \mathcal{L}_{aug}$  
& $7.08^\circ\pm0.12^\circ$
& $7.29^\circ\pm0.22^\circ$
& $7.35^\circ\pm0.09^\circ$
& $8.88^\circ\pm0.06^\circ$
& $7.65^\circ\textcolor{orange}{\ \blacktriangledown10.94\%}$
\\
$\mathcal{L}_{ori} + \mathcal{L}_{con}$   
& $7.34^\circ\pm0.10^\circ$
& $8.34^\circ\pm0.26^\circ$
& $7.76^\circ\pm0.05^\circ$
&$ 8.48^\circ\pm0.12^\circ$
& $7.98^\circ
\textcolor{orange}{\ \blacktriangledown7.10\%}$\\
$\mathcal{L}_{ori} + \mathcal{L}_{mmd}$ 
& $7.06^\circ\pm0.11^\circ$
& $7.19^\circ\pm0.15^\circ$
& $7.54^\circ\pm0.07^\circ$
& $9.53^\circ\pm0.30^\circ$
& $7.83^\circ
\textcolor{orange}{\ \blacktriangledown8.85\%}$\\
$\mathcal{L}_{ori} + \mathcal{L}_{aug} + \mathcal{L}_{con}$  
& $6.81^\circ\pm0.13^\circ$
& $6.85^\circ\pm0.26^\circ$
& $7.07^\circ\pm0.05^\circ$
& $8.34^\circ\pm0.15^\circ$
& $7.27^\circ
\textcolor{orange}{\ \blacktriangledown15.37\%}$
\\
$\mathcal{L}_{ori} + \mathcal{L}_{aug} + \mathcal{L}_{mmd}$  
& $6.64^\circ\pm0.14^\circ$
& $6.76^\circ\pm0.11^\circ$
& $7.39^\circ\pm0.06^\circ$
& $8.66^\circ\pm0.34^\circ$
& $7.36^\circ
\textcolor{orange}{\ \blacktriangledown14.32\%}$
\\
$\mathcal{L}_{ori} + \mathcal{L}_{con} + \mathcal{L}_{mmd}$ 
& $7.04^\circ\pm0.11^\circ$
& $7.09^\circ\pm0.10^\circ$
& $7.37^\circ\pm0.06^\circ$
& $9.41^\circ\pm0.26^\circ$
& $7.73^\circ \textcolor{orange}{\ \blacktriangledown10.01\%}$\\
$\mathcal{L}_{ori} + \mathcal{L}_{aug} + \mathcal{L}_{con} + \mathcal{L}_{mmd}$  & $6.35^\circ\pm0.10^\circ$
& $6.72^\circ\pm0.12^\circ$
& $6.96^\circ\pm0.07^\circ$
& $8.79^\circ\pm0.25^\circ$
& $7.21^\circ \textcolor{orange}{\ \blacktriangledown16.06\%}$\\
\bottomrule
\end{tabularx}
\end{table*}

\begin{table}[htbp]
\centering
\caption{Result of comparison with state-of-the-art gaze estimation and domain generalization methods}
\setlength{\tabcolsep}{2pt}
\begin{tabularx}{\linewidth}{lcccc}
\toprule
Method & $\mathcal{D}_E\rightarrow\mathcal{D}_M$ & $\mathcal{D}_E\rightarrow\mathcal{D}_D$ & $\mathcal{D}_G\rightarrow\mathcal{D}_M$ & $\mathcal{D}_G\rightarrow\mathcal{D}_D$ \\
\midrule
Full-Face \cite{zhang2017s} & $12.35^\circ$ & $30.15^\circ$ & $11.13^\circ$ & $14.42^\circ$ \\
CA-Net \cite{cheng2020coarse} & - & - & $27.13^\circ$& $31.41^\circ$\\
RT-Gene \cite{fischer2018rt} & - & - & $21.81^\circ$& $38.60^\circ$\\
Dilated-Net \cite{chen2018appearance} & - & - & $18.45^\circ$& $23.88^\circ$ \\
ADL* \cite{kellnhofer2019gaze360} & $7.23^\circ$& $8.02^\circ$& $11.36^\circ$& $11.86^\circ$\\
\midrule
PureGaze \cite{cheng2022puregaze} & $7.08^\circ$& $7.48^\circ$& $9.28^\circ$& $9.32^\circ$\\
Gaze-Consistent \cite{xu2023learning} & $6.50^\circ$& $7.44^\circ$& $7.55^\circ$& $9.03^\circ$\\
\midrule
Baseline & $8.02^\circ$& $9.11^\circ$& $8.04^\circ$& $9.20^\circ$ \\
\textbf{Gaze-BAR} & $\textbf{6.35}\boldsymbol{^\circ}$ & $\textbf{6.72}\boldsymbol{^\circ}$ & $\textbf{6.96}\boldsymbol{^\circ}$ & $\textbf{8.79}\boldsymbol{^\circ}$ \\
\bottomrule
\end{tabularx}
\end{table}

\subsection{Data Preparation and Evaluation Metric}

In this letter, we validate the efficacy of our model across four distinct datasets: ETH-XGaze ($\mathcal{D}_E$) \cite{zhang2020eth}, Gaze360 ($\mathcal{D}_G$) \cite{kellnhofer2019gaze360}, MPIIGaze ($\mathcal{D}_M$) \cite{zhang2017mpiigaze}, and EyeDiap ($\mathcal{D}_D$) \cite{funes2014eyediap}. 
ETH-XGaze encompasses 110 subjects, contributing a total of 1.1M images, with a training set provided for 80 subjects. We partition 5 subjects from this pool for validation purposes. Gaze360 comprises 238 subjects, yielding 172K images after excluding those devoid of eye-related content. Following the protocol proposed by Sugano et al. \cite{sugano2014learning}, we process MPIIGaze to extract 45K images from 15 subjects. EyeDiap offers 94 video clips from 16 subjects, from which we sample one image every 15 frames for analysis.

Following~\cite{cheng2022puregaze, xu2023learning}, we designate ETH-XGaze and Gaze360 as training sets, while MPIIGaze and EyeDiap are reserved for model evaluation. Consequently, four cross-dataset tasks are investigated in our experiments: $\mathcal{D}_E\rightarrow\mathcal{D}_M$, $\mathcal{D}_E\rightarrow\mathcal{D}_D$, $\mathcal{D}_G\rightarrow\mathcal{D}_M$, and $\mathcal{D}_G\rightarrow\mathcal{D}_D$.

In all experiments of this paper, we leverage the angular error in Eq. (2) as the evaluation metric. Smaller angular error means better inference accuracy.  

\subsection{Implementation Details}

The experiments were conducted using the PyTorch framework and on a single RTX 4080 GPU. 
Our model was trained under the batch size of 64 and the Adam optimizer with a learning rate of $1 \times 10^{-4}$. 
We resize and normalize the input images to $224 \times 224$ and $[0,1]$. The model was trained for 10 epochs on ETH-XGaze and 100 epochs on Gaze360.

\subsection{Comparison Methods}
\subsubsection{Baseline}
The baseline model is the original single-branch gaze estimation model consisting of a ResNet-18 pre-trained on ImageNet as the backbone and an MLP module.

\subsubsection{State-of-the-art methods}
\begin{itemize}
    \item Gaze estimation methods of Full-Face \cite{zhang2017s}, CA-Net \cite{cheng2020coarse}, RT-Gene \cite{fischer2018rt}, Dilated-Net \cite{chen2018appearance}, and ADL* \cite{kellnhofer2019gaze360}.
    \item Domain generalization methods for gaze estimation including PureGaze \cite{cheng2022puregaze} and Gaze-Consistent \cite{xu2023learning}.
\end{itemize}

\subsection{Ablation Studies}

This section presents ablation studies aimed at dissecting the contributions of each component within our approach. We assess the impact of removing $\mathcal{L}_{aug}$, $ \mathcal{L}_{mmd}$, or $\mathcal{L}_{con}$, comparing these variations with both the baseline model and our complete methodology.

The results, summarized in Table I, indicate that combining the original loss (\(\mathcal{L}_{ori}\)) with the additional loss terms leads to significant improvements in gaze estimation accuracy. Notably, the most substantial improvement is observed when all loss components are integrated, yielding a reduction of $16.06\%$ in average angular error compared to the baseline. These findings underscore the efficacy of the proposed multi-loss framework in enhancing gaze estimation across diverse datasets.

\subsection{State-of-the-Art Comparison}

In this section, we compare the performance of our proposed domain generalization method against SOTA gaze estimation methods. Table II presents the angular error results for various methods over four cross-dataset tasks.

Our method consistently outperforms existing SOTA methods in all cross-dataset tasks, achieving the lowest angular error for every cross-domain task. This underscores the advanced generalization ability of our approaches to handle variations in environmental factors and facial attributes. 

\subsection{Plug Into Existing Gaze Estimation Models}

To substantiate the portability of our approach, we complement our analysis by integrating existing gaze estimation models into our comparative evaluation framework. Table III showcases the performance enhancements achieved by plugging our proposed method, BAR (Branch-out Auxiliary Regularization), into existing gaze estimation models over four cross-dataset tasks. 

\begin{table}[htbp]
\centering
\caption{ Performance of our Branch-out Auxiliary Regularization plugged to existing models}
\setlength{\tabcolsep}{1.3pt}
\renewcommand{\arraystretch}{1.1}
\begin{tabularx}{\linewidth}{lcccc}
\toprule
Method & $\mathcal{D}_E\rightarrow\mathcal{D}_M$ & $\mathcal{D}_E\rightarrow\mathcal{D}_D$ & $\mathcal{D}_G\rightarrow\mathcal{D}_M$ & $\mathcal{D}_G\rightarrow\mathcal{D}_D$ \\
\midrule
Full-Face \cite{zhang2017s} & $12.35^\circ$& $30.15^\circ$&$ 11.13^\circ$& $14.42^\circ$ \\
Full-Face \cite{zhang2017s} + BAR & $9.04^\circ$& $9.45^\circ$& $8.70^\circ$& $11.33^\circ$ \\ 
\midrule
Dilated-Net \cite{chen2018appearance} & - & - & $18.45^\circ$& $23.88^\circ$\\
Dilated-Net \cite{chen2018appearance} + BAR& - & - & $14.22^\circ$& $16.55^\circ$ \\
\midrule
Baseline & $8.02^\circ$& $9.11^\circ$& $8.04^\circ$& $9.20^\circ$\\
Baseline + BAR& $\textbf{6.35}\boldsymbol{^{\circ}}$ & $\textbf{6.72}\boldsymbol{^\circ}$ & $\textbf{6.96}\boldsymbol{^\circ}$ & $\textbf{8.79}\boldsymbol{^\circ}$ \\
\bottomrule
\end{tabularx}
\end{table}

Our results indicate consistent improvements across all evaluated methods when integrated with the BAR framework. Notably, the Full-Face and Dilated-Net models witness a significant reduction in angular error, showcasing the efficacy of our approach in enhancing the generalization performance of existing gaze estimation architectures.

\begin{figure}[htbp]
\centering
\subfloat[Baseline: $\mathcal{D}_E\rightarrow\mathcal{D}_M$]{\includegraphics[width=0.23\textwidth]{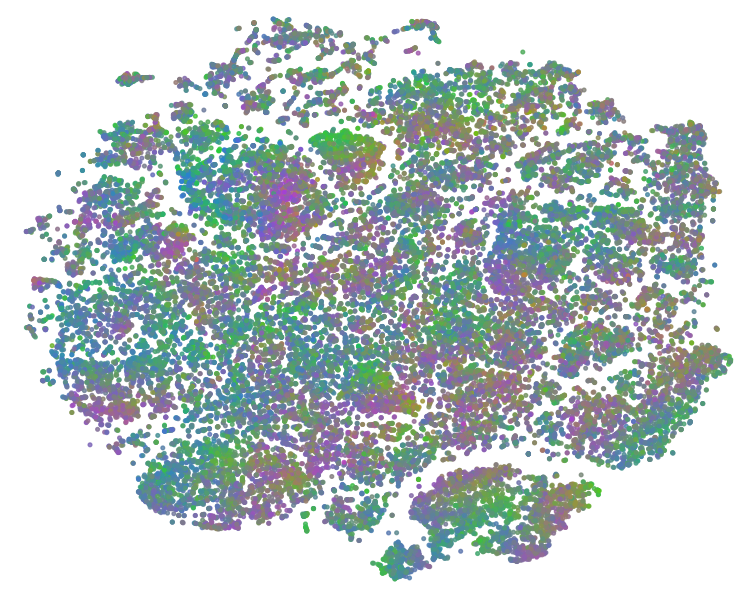}}
\subfloat[Gaze-BAR: $\mathcal{D}_E\rightarrow\mathcal{D}_M$]{\includegraphics[width=0.23\textwidth]{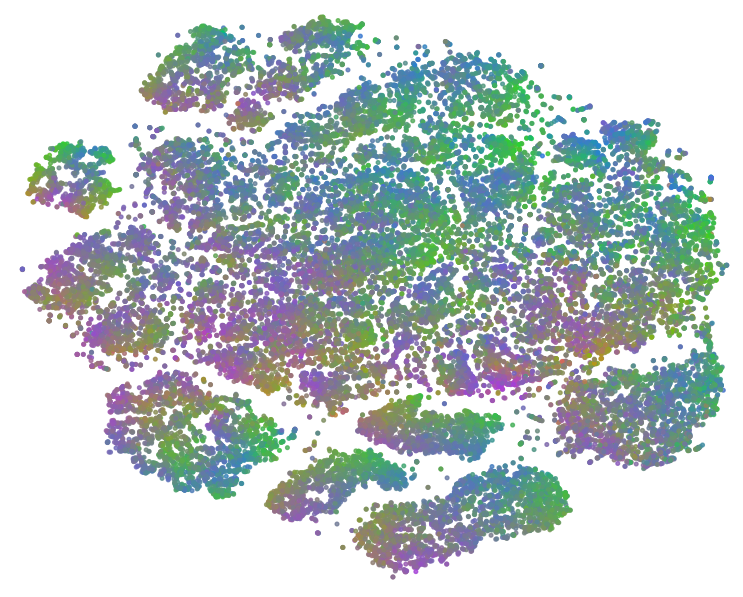}}
\caption[]{t-SNE visualization of the extracted features. Similar colors represent similar gaze directions. Each dot denotes a sample from the test set.}\label{fig:vis}
\end{figure}

\subsection{Visualization of Gaze Feature}
To compare and analyze the features acquired by the baseline model and our model, we employ t-SNE \cite{van2008visualizing} to visualize the feature distribution of task $\mathcal{D}_E\rightarrow\mathcal{D}_M$. The results are depicted in Fig.~\ref{fig:vis}, where similar colors represent similar gaze directions. The feature distribution of the baseline model demonstrates that there is no discernible correlation between gaze features and gaze direction. However, in our methodology, a noticeable transition in color from green to purple is observed from the upper right to the lower left corner, validating our model's capability to extract gaze representations consistent with the gaze direction.

\section{Conclusion}
In this letter, we introduce branch-out auxiliary regularization, a straightforward and flexible approach to enhance domain generalization in gaze estimation models without
requiring direct access to target domain data. Our method leverages data augmentation to address environmental variations and incorporates a contrast branch for aligning gaze directions with positive samples from the source domain, thereby learning consistent gaze features. Comprehensive experimental evaluations on four cross-dataset tasks demonstrate the superiority of our approach.

\balance

\bibliographystyle{IEEEtran}
\bibliography{ref} 

\end{document}